\documentclass[conference,letterpaper]{IEEEtran}
\usepackage[letterpaper, left=0.75in, right=0.75in, bottom=0.75in, top=0.75in]{geometry}
\def\IEEEtitletopspace{18pt}
\IEEEoverridecommandlockouts
\usepackage{array}
\usepackage{hyperref}
\usepackage{cite}
\usepackage{amsmath,amssymb,amsfonts}
\usepackage{algorithmic}
\usepackage{comment}
\usepackage{graphicx}
\usepackage{adjustbox}
\usepackage{textcomp}
\usepackage{breqn}
\usepackage{amsmath}
\usepackage{xcolor}
\usepackage{booktabs}
\usepackage{xcolor}
\usepackage{subcaption}
\usepackage{caption}
\newcount\Comments  
\usepackage{color}
\definecolor{darkgreen}{rgb}{0,0.5,0}
\definecolor{purple}{rgb}{1,0,1}
\definecolor{teal}{rgb}{0,0.4627,0.5804}
\newcommand{\kibitz}[2]{\ifnum\Comments=1\textcolor{#1}{#2}\fi}

\def\BibTeX{{\rm B\kern-.05em{\sc i\kern-.025em b}\kern-.08em
    T\kern-.1667em\lower.7ex\hbox{E}\kern-.125emX}}
\begin{document}

\title{Airport Delay Prediction with Temporal Fusion Transformers\\

%


\thanks{
$^{*}$K. Liu (corresponding author) is with the University of California at Berkeley, Berkeley, CA 94702, USA. 






}
}

\author{Ke Liu$^{*}$, Kaijing Ding, Xi Cheng, Guanhao Xu, Xin Hu, Tong Liu, Siyuan Feng, Binze Cai, \\ Jianan Chen, Hui Lin, Jilin Song, Chen Zhu}

\maketitle

\begin{abstract}

Since flight delay hurts passengers, airlines, and airports, its prediction becomes crucial for the decision-making of all stakeholders in the aviation industry and thus has been attempted by various previous research. However, previous delay predictions are often categorical and at a highly aggregated level. To improve that, this study proposes to apply the novel Temporal Fusion Transformer model and predict numerical airport arrival delays at quarter hour level for U.S. top 30 airports. Inputs to our model include airport demand and capacity forecasts, historic airport operation efficiency information, airport wind and visibility conditions, as well as enroute weather and traffic conditions. The results show that our model achieves satisfactory performance measured by small prediction errors on the test set. In addition, the interpretability analysis of the model outputs identifies the important input factors for delay prediction.

\end{abstract}


\section{Introduction}
The aviation industry, despite experiencing a significant downturn in air traffic demand during the global pandemic, is now on a path to recovery. Projections indicate an annual growth in traffic of 1.5\% to 3.8\% over the next two decades (\cite{IATA}). Such growth, while promising, forecasts a burgeoning gap between demand and capacity at airports, potentially exacerbating flight delays—an outcome that significantly impacts passenger satisfaction, airline operating costs, and environmental sustainability.

To mitigate these challenges, it is increasingly vital for aviation authorities to develop robust mechanisms for predicting flight delays and to establish more efficient traffic management initiatives (TMIs). The literature is replete with studies aimed at forecasting flight delays using a variety of methodologies (\cite{review}). One of the seminal works in this field by (\cite{Kim2016}) employed deep learning techniques, specifically the Long Short Term Memory (LSTM) model, to analyze day-to-day sequences of departure and arrival delays at a single airport, successfully predicting delay classes based on predefined thresholds. Another noteworthy study by (\cite{Vandal2018}) utilized the LSTM model to predict aggregated daily delays for 123 U.S. airports, incorporating Monte Carlo Dropout techniques to refine parameter variance estimates. However, most existing studies in flight delay prediction focus on binary outcomes (delayed or not) or on categorizing delays into broad classes. Furthermore, these predictions often pertain to highly aggregated levels, such as daily delay forecasts.

Building on recent advancements, a cutting-edge study (\cite{mitre2021}) demonstrated the application of transformers to predict airport delays at the quarter-hour level across several regional airports. This has paved the way for the adoption of more sophisticated models capable of tackling aviation challenges with greater precision.

In this paper, we propose a novel approach to predict the specific numerical values of airport arrival delays at a more granular level—specifically, every quarter-hour over a strategic horizon of up to four hours. We focus on the top 30 U.S. airports, incorporating variables such as airport demand and capacity forecasts, historical operational efficiency, and local weather conditions. To achieve this, we will deploy the Temporal Fusion Transformer (TFT), an attention-based deep neural network model designed for multi-horizon forecasting. The TFT model has demonstrated superior performance over other forecasting techniques like DeepAR, ARIMA, and traditional LSTM Seq2Seq across various datasets (\cite{google}), validating its effectiveness for time-series data analysis.

The paper is structured as follows: Section 2 provides an overview of the datasets utilized and the preprocessing of input variables. Section 3 delves into the modeling details, while Section 4 presents and discusses the results. We conclude our study in Section 5 with a summary of our findings and insights.

\nocite{bloem2015ground}
\nocite{ganesan2010predicting}
\nocite{mukherjee2014predicting}
\nocite{george2015reinforcement}
\nocite{cox2016ground}
\nocite{zhu2017communication}
\nocite{kalliguddi2017predictive}
\nocite{yilmaz2021deep}
\nocite{deshpande2012impact}

\section{Data}
\subsection{Data Sources}

For this study, we focused on the top 30 busiest airports in the U.S. during the year 2016, spanning from January 1st to December 31st. To conduct a comprehensive analysis, we collected and integrated three key datasets from the Federal Aviation Administration's (FAA) Aviation System Performance Metrics (ASPM) and the Integrated Surface Database (ISD). Prior to integration, each dataset underwent rigorous cleaning, filtering, and time zone normalization to UTC to ensure data consistency and accuracy. The resulting master database is structured on a quarter-hourly basis for each airport, amounting to a total of 1,054,080 data points (30 airports × 366 days × 24 hours per day × 4 quarters per hour).

The datasets utilized include:
\begin{enumerate}
\item \textbf{FAA ASPM Flight Level Data:} This dataset provides detailed records for each flight, including flight plans, scheduled and actual times, and Estimated Departure Clearance Time (EDCT) for flights arriving at 77 major U.S. airports. It serves as a crucial source for analyzing flight-specific delay patterns and operational efficiency.

\item \textbf{FAA ASPM Airport Quarter-Hour Data:} Contains comprehensive data on operational conditions at 15-minute intervals. This dataset includes information on airport capacity, runway configurations, and terminal weather conditions, essential for understanding the operational dynamics that influence flight delays at the airport level.

\item \textbf{Global Hourly – Integrated Surface Database (ISD):} Comprises hourly weather records from 2,330 surface stations across the U.S., providing extensive meteorological data crucial for correlating weather conditions with flight delays. The details of the stations and their data coverage are documented in Appendix 1(c).
\end{enumerate}

This robust integration of flight, airport operational, and weather data provides a solid foundation for developing predictive models aimed at forecasting airport delays with high precision. By harnessing detailed historical data, our approach seeks to uncover nuanced relationships between airport operations and delay occurrences, enabling more accurate and timely predictions that could significantly enhance traffic management and operational planning at major U.S. airports.

This section outlines the input variables for airport-level delay prediction and describes their processing from raw datasets. The primary data sources are from the FAA's ASPM airport quarter-hour dataset. Inputs include:

Airport ID: Captures systematic variations in traffic management efficiency and congestion at different airports.
Time Index, Month, Local Hour, and Day of the Week: Helps capture general traffic volume and flight operation patterns.
Scheduled Flight Departure and Arrival Counts: Set 2-6 months in advance of the actual flight day.
Airport Capacity Forecasts for Departures and Arrivals: Predictive data on expected airport throughput.
Observed Airport Arrival and Departure Throughput: Actual counts of flights handled by the airport.
Observed Airport Demand: The number of flights scheduled to arrive and depart.
On-Time Percentage for Arrivals and Departures: Statistics on the punctuality of flights.
Average Arrival/Departure Delays: Prediction variable, smoothed by moving average over adjacent three quarter hours to mitigate extreme values.
Additionally, to assess congestion effects, we calculate cumulative queuing delays for arrivals and departures using a deterministic queuing model. This model operates with quarter-hour intervals, counting demand based on scheduled 'wheels-on' times and throughput based on actual 'wheels-on' times. If a flight lands earlier than scheduled, it is only counted in the earlier interval. We then compute the cumulative actual arrivals and demands for each quarter-hour, where the area between the arrival and demand curves indicates total queuing delays for that day.

For operational variables, we include:
\begin{itemize}
     \item Enroute Traffic Density: This considers the traffic managed by terminal air traffic controllers, who are responsible for aircraft entering and exiting the airport and ensuring safe separation over the busy surrounding airspace. The U.S. airspace is divided into 0.25° cells, extending from 25°N to 50°N latitude and 66°W to 125°W longitude. Using distance-based interpolation along the great circle route, we track en-route locations at quarter-hour intervals, aggregating traffic density across the grid. An example of this grid system is shown in Appendix 1(a).
 \item Convective Weather Factors: Thunderstorms, crosswinds, tailwinds, ceiling, and visibility at airports are considered. Thunderstorms, in particular, can drastically reduce airport capacity. We resample weather data hourly and apply 2D grid interpolation to assign weights to grid cells for convective weather, creating hourly weather matrices for the airspace grid (see \cite{liu2023airborne} for detailed feature engineering).
 \end{itemize}
This detailed input processing approach allows for a nuanced understanding of factors influencing airport delays, ensuring our predictive models are both comprehensive and precise.

\begin{figure*} [ht]
    \centering
    \includegraphics[width=0.75\linewidth]{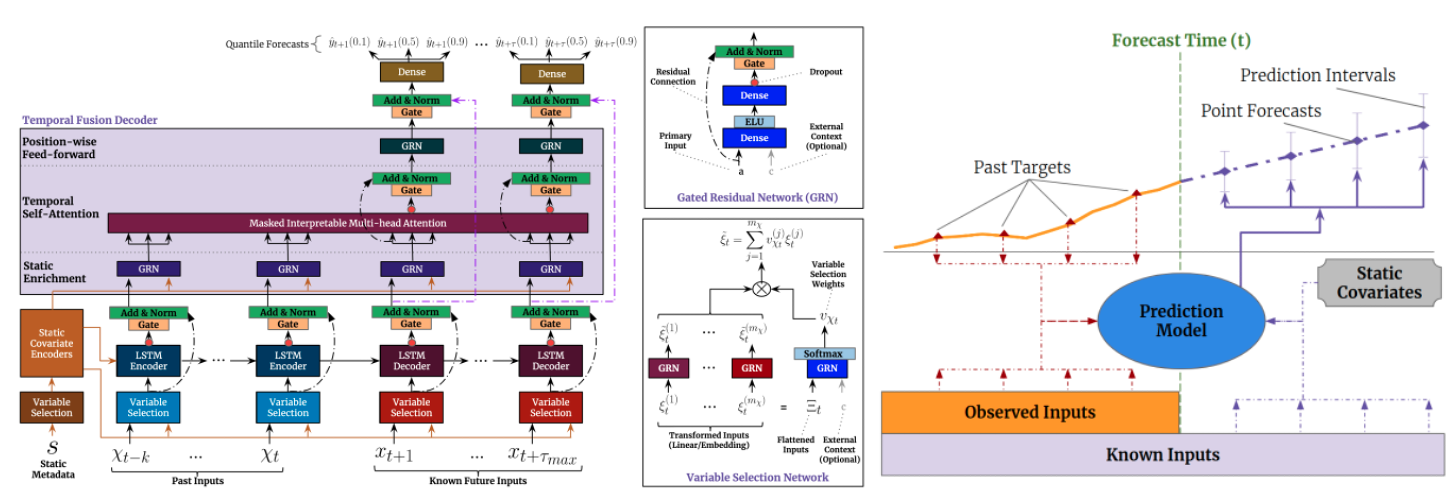}
    \caption{Left_TFT Architecture (\cite{lim2019}; Right:Illustration of multi-horizon forecasting(\cite{lim2019})}
    \label{fig:mae}
\end{figure*}


\section{Modeling}
\subsection{Temporal Fusion Transformer}
Temporal Fusion Transformer (TFT) is an attention-based architecture which combines high-performance multi-horizon forecasting with interpretable insights into temporal dynamics, first proposed by \cite{lim2019}. It integrates the mechanisms of several other neural architectures we learned in class, for instance LSTM layers and the attention heads used in Transformers. The major components of TFT include gating mechanisms, variable selection networks, static covariate encoders, temporal processing, and prediction intervals via quantile forecasts. Details can be found in \cite{lim2019} and the general architecture is shown in Fig. \ref{fig:mae}.

Compared with the classic transformer model, it has several advantages and novelties, which makes it a good fit for our work. First, the input features to TFT can be of three types: i) temporal data with known inputs into the future, e.g., the future airport demands; ii) temporal data known only up to the present, e.g., the historical airport delays; and iii) exogenous categorical/static variables, also known as time-invariant features, e.g. airport IDs. Second,  it supports multi-step predictions. This characteristic facilitate our prediction of airport arrival delays for different quarter hours in the future. Furthermore, TFT also outputs prediction intervals, by using the quantile loss function. Therefore, TFT can provide range estimates rather than a single point estimate, which will offer air traffic controllers more information when they are making decisions. In addition, TFT has good interpretability, which can help identify the key contributions to airport delays. Together with its advantages in high performance and available open source implementations, all these strengths makes it a preferrable model for us to implement delay prediction.

\subsection{Problem Formulation}

The goal of this study is to predict arrival delays for the U.S. Top 30 airports over a strategic time horizon. In a generic form, it can be formulated as a multi-variate, multi-step, time series forecasting problem. Here we determined the time lag variable, or the look-back time of what has happened as 2 hours, or 8 time steps (since our data is in quarter hours); and the maximum look-ahead time as 4 hours, or 16 time steps, which is determined by the data updating frequency as well as prediction needs. Fig. \ref{fig:mae} further illustrates the relationship of inputs and outputs in
a given time horizon. The inputs and the outputs of our TFT model are prepared from the sources specified in Section 2. Table \ref{table:inputs}  lists the detailed inputs to TFT for delay prediction.

\begin{table}[h]
    \centering\small
    \setlength{\leftmargini}{0.4cm}
    \caption{Input Variables Description}\label{table:inputs}
\begin{tabular}{ | m{2cm}| m{5.5cm}| } 
 \hline
 Variable Types & Variables\\ 
 \hline
 Static Covariates & 
 Airport ID, Month, Local hour, Day of the week\\
 \hline
 Past-Observed Inputs &
 \begin{itemize} 
 \setlength\itemsep{0.05em}
    \item Actual arrival and departure counts
    \item Reported number of aircrafts intending to arrive and departure for the period
    \item Average arrival delays (based on flight plan) and departure delays
    \item Arrival and departure on-time percentage statistics
    \item Cumulative arrival and departure queuing delays
 \end{itemize}\\ 
 \hline
 Apriori-Known Inputs & 
 \begin{itemize}
 \setlength\itemsep{0.05em}
     \item Scheduled quarter-hourly arrival and departure demand
     \item Quarter-hourly airport arrival and departure capacity
     \item Airport visibility and ceiling conditions
     \item Airport headwind and tailwind conditions of arrival and departure runways
     \item Enroute convective weather
     \item Enroute traffic densities
 \end{itemize}\\
 \hline
\end{tabular}
\end{table}

\begin{figure*} [ht]
    \centering
    \includegraphics[width=1\linewidth]{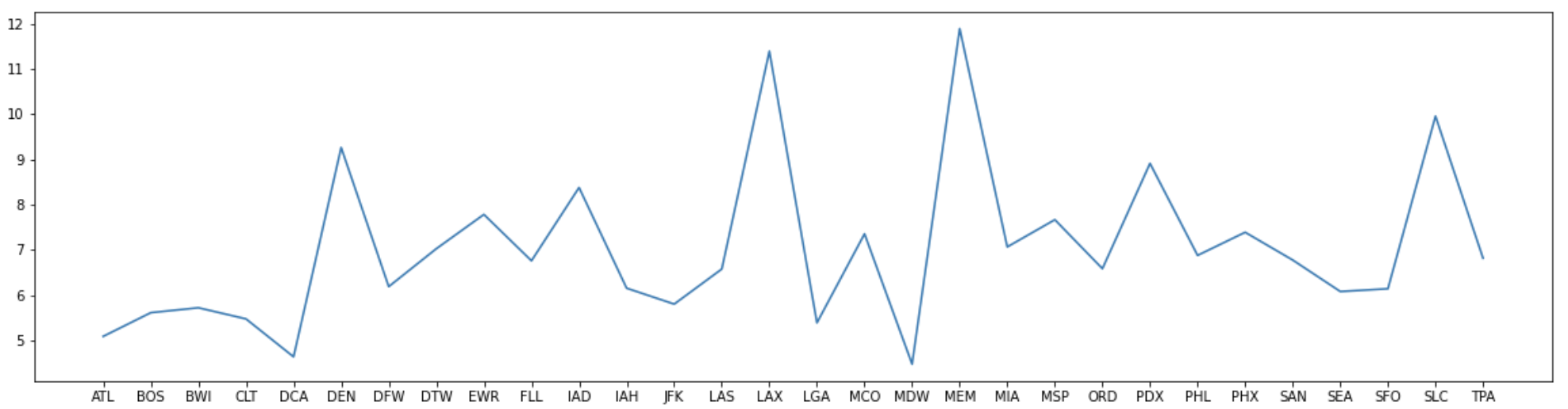}
    \caption{Model Performance for 30 Airports Measured by MAE(min)}
    \label{fig:mae}
\end{figure*}

\begin{figure*} [ht]
    \centering
    \includegraphics[width=1\linewidth]{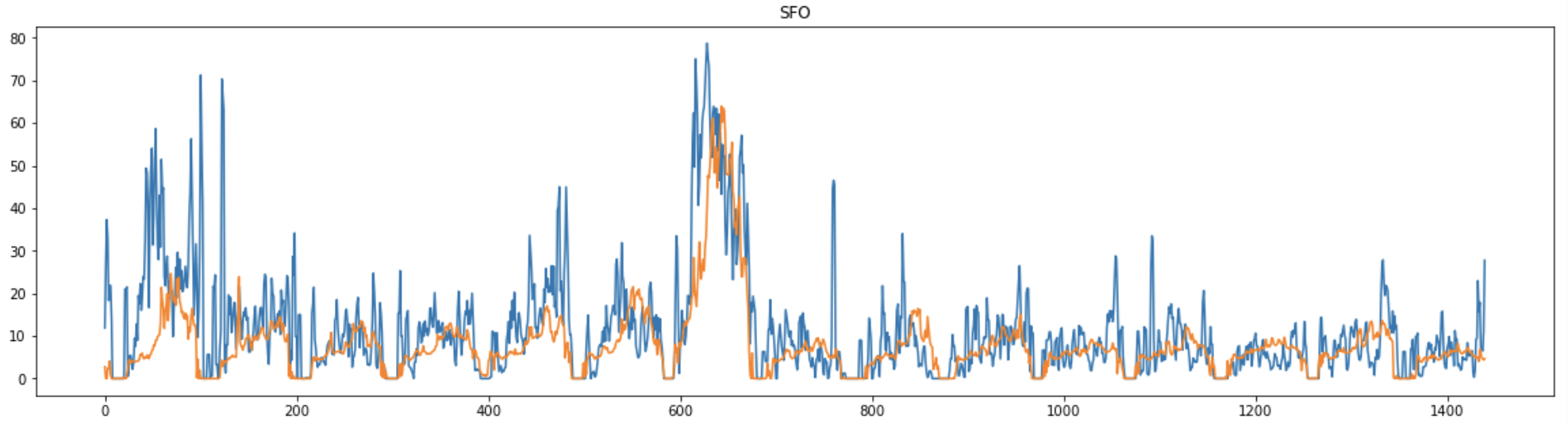}
    \caption{Model prediction of arrival delays(min) at SFO for selected testing days}
    \label{fig:sfo}
\end{figure*}

With these inputs, we use the TemporalFusionTransformer model given by PyTorch Forecasting package to train our flight delay forecast model. The model is trained and validated on the first eleven and a half months and the last 15 days of the year are set aside as test set. The hyperparameters are finely tuned, including but not limited to size of the hidden layers, dropout rate, attention head size, and learning rate; and the parameters are learned. 

\section{Results}

\subsection{Delay Prediction Results}

Fig. \ref{fig:mae} shows the model performance for the 30 airports on the testing dataset of the last 15 days of 2016, measured by mean absolute error(MAE). The performance of the model varies among different airports, with MAE ranging from 5 minutes to 12 minutes. In general, airports with higher delays have a higher MAE.

\begin{figure*} [ht]
    \centering
    \includegraphics[width=0.5\linewidth]{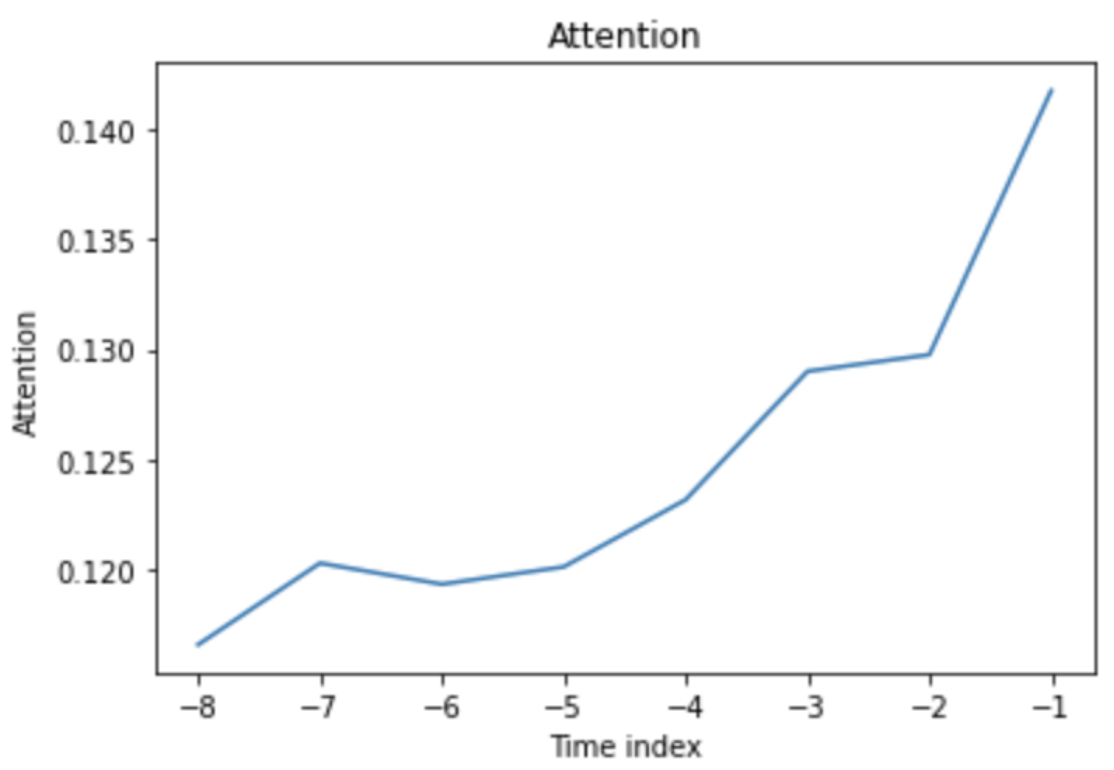}
    \caption{Attention score by time index}
    \label{fig:attention}
\end{figure*}

To understand the temporal differences of the prediction performance, Fig. \ref{fig:sfo} show the prediction
results of arrival delays at SFO for selected testing days. The comparison of actual (blue line) with predicted (orange line) values 
demonstrates that the TFT model can capture most upward and downward trends of delays.

\begin{figure*} [ht]
\centering
\hspace*{-0in}
\begin{subfigure}{.5\textwidth}
  \centering
  \includegraphics[width=0.7\linewidth]{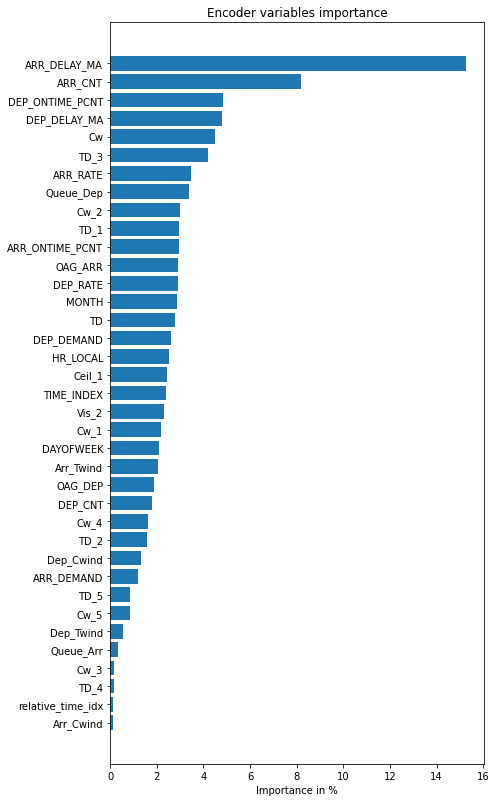}
  \label{fig:exp1}
\end{subfigure}%
\hspace*{0in}
\begin{subfigure}{.5\textwidth}
  \centering
  \includegraphics[width=0.9\linewidth]{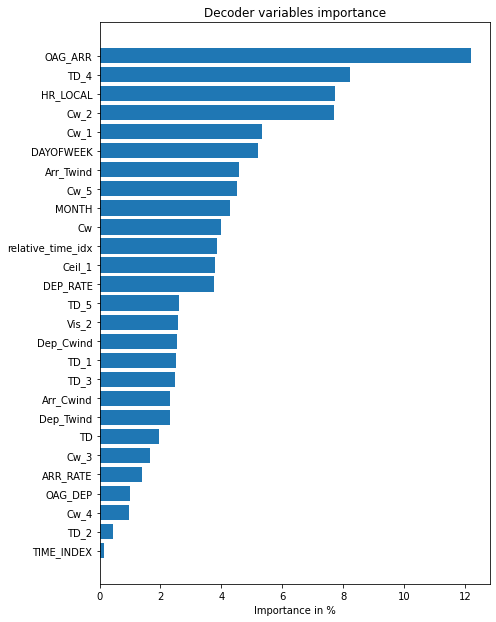}
  \label{fig:exp2}
\end{subfigure}
\caption{Importance of variables}
\label{fig:importance}
\end{figure*}
However, the model still does not make predictions that match the ground truth perfectly. Possible causes are also examined. First, although we have moved average the delay data, it still has a huge variation within adjacent quarter hours. This unsmoothness characteristic of the target variable makes it difficult to predict. Second, this study applies the model to all U.S. core 30 airports while the different airports may be influenced by the inputs differently. For example, SFO may be more impacted by bad weathers than LAX is, which problem, however, cannot be handled by the setting of this model. Third, there are other factors influencing flight delays that are not captured by the dataset we collect. For example, we do not have access to traffic management data like Ground Delay Program implementation or Airspace Flow Program implementation, which can better capture the traffic impact of other airports and enroute airspace on the study airport.  Last but not least, the training and test dataset may not be of the same distribution. The second half of December (test set) is likely to have more severe weather days than the rest of the year. Due to the imbalanced nature of the data, the model would perform better (in terms of error minimization) on the non-severe weather days.

\subsection{Model Interpretation}
TFT has an advantage in interpretability of time dynamics. For example, Fig. \ref{fig:attention} shows the attention score of each time index. A higher attention score means a bigger contribution from that time step to the prediction outputs. Since the look-back time of our model is set to 2 hours, or 8 time steps, the time indices are from -8 to -1. As expected, the closer time steps receive higher scores and thus make a bigger contribution.

Furthermore, the TFT model is also interpretable in terms of the importance of each input variable for the response variables. Based on Fig. \ref{fig:importance}, the relative importance of the input variables for delay prediction can be identified. For the encoder inputs, historic arrival delays and arrival counts are top 2 factors. For the decoder input variables, scheduled arrival demands, enroute traffic density, local hour, enroute convective weather are influential factors.

\section{Summary}
In this study, we applied Temporal Fusion Transformer models to the prediction of average flight arrival delay of U.S. core 30 airports at quarter hour interval with the maximum look-ahead time as 4 hours(or 16 time steps). The inputs of model comes from a wide range of data and are processed to involve the information of airport operation conditions (capacity, schedule flights, demand), terminal and nearby enroute congestion level, airport weather conditions and nearby enroute thunderstorm. The TFT model handle well with these heterogeneous inputs. The performance of the model, though varies among airports, is acceptable in that it captures most upward and downward trends of delays. Future work can be carried out to process the convective weather information better, to include the enroute wind information and most importantly, to incorporate the TMIs (such as mile-in-trail, and ground delay program).

\bibliographystyle{IEEEtran}
\bibliography{refs.bib,references.bib}

\begin{thebibliography}{10}
\providecommand{\url}[1]{#1}
\csname url@samestyle\endcsname
\providecommand{\newblock}{\relax}
\providecommand{\bibinfo}[2]{#2}
\providecommand{\BIBentrySTDinterwordspacing}{\spaceskip=0pt\relax}
\providecommand{\BIBentryALTinterwordstretchfactor}{4}
\providecommand{\BIBentryALTinterwordspacing}{\spaceskip=\fontdimen2\font plus
\BIBentryALTinterwordstretchfactor\fontdimen3\font minus \fontdimen4\font\relax}
\providecommand{\BIBforeignlanguage}[2]{{%
\expandafter\ifx\csname l@#1\endcsname\relax
\typeout{** WARNING: IEEEtran.bst: No hyphenation pattern has been}%
\typeout{** loaded for the language `#1'. Using the pattern for}%
\typeout{** the default language instead.}%
\else
\language=\csname l@#1\endcsname
\fi
#2}}
\providecommand{\BIBdecl}{\relax}
\BIBdecl

\bibitem{IATA}
\BIBentryALTinterwordspacing
IATA, ``20 year passenger forecast,'' 2019. [Online]. Available: \url{https://www.iata.org/en/publications/store/20-year-passenger-forecast/}
\BIBentrySTDinterwordspacing

\bibitem{review}
\BIBentryALTinterwordspacing
A.~Sternberg, J.~Soares, D.~Carvalho, and E.~Ogasawara, ``A review on flight delay prediction,'' 2017. [Online]. Available: \url{https://arxiv.org/abs/1703.06118}
\BIBentrySTDinterwordspacing

\bibitem{Kim2016}
Y.~J. Kim, S.~Choi, S.~Briceno, and D.~Mavris, ``A deep learning approach to flight delay prediction,'' in \emph{2016 IEEE/AIAA 35th Digital Avionics Systems Conference (DASC)}.\hskip 1em plus 0.5em minus 0.4em\relax IEEE, 2016, pp. 1--6.

\bibitem{Vandal2018}
T.~Vandal, M.~Livingston, C.~Piho, and S.~Zimmerman, ``Prediction and uncertainty quantification of daily airport flight delays,'' in \emph{Proceedings of The 4th International Conference on Predictive Applications and APIs}.\hskip 1em plus 0.5em minus 0.4em\relax PMLR, 2018, pp. 82:45--51.

\bibitem{mitre2021}
\BIBentryALTinterwordspacing
L.~Wang, A.~Tien, and J.~Chou, ``Multi-airport delay prediction with transformers,'' 2021. [Online]. Available: \url{https://arxiv.org/abs/1912.09363}
\BIBentrySTDinterwordspacing

\bibitem{google}
\BIBentryALTinterwordspacing
S.~O. Arik and T.~Pfister, ``Interpretable deep learning for time series forecasting,'' 2021. [Online]. Available: \url{https://ai.googleblog.com/2021/12/interpretable-deep-learning-for-time.html}
\BIBentrySTDinterwordspacing

\bibitem{bloem2015ground}
M.~Bloem and N.~Bambos, ``Ground delay program analytics with behavioral cloning and inverse reinforcement learning,'' \emph{Journal of Aerospace Information Systems}, vol.~12, no.~3, pp. 299--313, 2015.

\bibitem{ganesan2010predicting}
R.~Ganesan, P.~Balakrishna, and L.~Sherry, ``Predicting aircraft taxi-out times,'' Jul.~22 2010, uS Patent App. 12/688,131.

\bibitem{mukherjee2014predicting}
A.~Mukherjee, S.~R. Grabbe, and B.~Sridhar, ``Predicting ground delay program at an airport based on meteorological conditions,'' in \emph{14th AIAA aviation technology, integration, and operations conference}, 2014, p. 2713.

\bibitem{george2015reinforcement}
E.~George and S.~S. Khan, ``Reinforcement learning for taxi-out time prediction: An improved q-learning approach,'' in \emph{2015 International Conference on Computing and Network Communications (CoCoNet)}.\hskip 1em plus 0.5em minus 0.4em\relax IEEE, 2015, pp. 757--764.

\bibitem{cox2016ground}
J.~Cox and M.~J. Kochenderfer, ``Ground delay program planning using markov decision processes,'' \emph{Journal of Aerospace Information Systems}, vol.~13, no.~3, pp. 134--142, 2016.

\bibitem{zhu2017communication}
L.~Zhu, Y.~He, F.~R. Yu, B.~Ning, T.~Tang, and N.~Zhao, ``Communication-based train control system performance optimization using deep reinforcement learning,'' \emph{IEEE Transactions on Vehicular Technology}, vol.~66, no.~12, pp. 10\,705--10\,717, 2017.

\bibitem{kalliguddi2017predictive}
A.~M. Kalliguddi and A.~K. Leboulluec, ``Predictive modeling of aircraft flight delay,'' \emph{Universal Journal of Management}, vol.~5, no.~10, pp. 485--491, 2017.

\bibitem{yilmaz2021deep}
E.~Yilmaz, O.~Sanni, M.~Kotwicz~Herniczek, and B.~German, ``Deep reinforcement learning approach to air traffic optimization using the muzero algorithm,'' in \emph{AIAA AVIATION 2021 FORUM}, 2021, p. 2377.

\bibitem{deshpande2012impact}
V.~Deshpande and M.~Ar{\i}kan, ``The impact of airline flight schedules on flight delays,'' \emph{Manufacturing \& Service Operations Management}, vol.~14, no.~3, pp. 423--440, 2012.

\bibitem{liu2023airborne}
K.~Liu, Z.~Zheng, B.~Zou, and M.~Hansen, ``Airborne flight time: A comparative analysis between the us and china,'' \emph{Journal of Air Transport Management}, vol. 107, p. 102341, 2023.

\bibitem{lim2019}
\BIBentryALTinterwordspacing
B.~Lim, S.~O. Arik, N.~Loeff, and T.~Pfister, ``Temporal fusion transformers for interpretable multi-horizon time series forecasting,'' 2019. [Online]. Available: \url{https://arxiv.org/abs/1912.09363}
\BIBentrySTDinterwordspacing

\end{thebibliography}

\vspace{12pt}

\end{document}